\documentclass[11pt,a4paper]{article}
\usepackage[nohyperref]{acl2018}
\usepackage{times}
\usepackage{latexsym}
\usepackage{graphicx}
\usepackage{booktabs}
\usepackage{multirow}
\usepackage{fixltx2e}
\usepackage{algorithm} 
\usepackage{algpseudocode}

\usepackage{url}

\aclfinalcopy 
\def\aclpaperid{1555} 

\title{Multimodal Affective Analysis Using Hierarchical Attention Strategy with Word-Level Alignment}

\newcommand*\samethanks[1][\value{footnote}]{\footnotemark[#1]}
\author{Yue Gu, Kangning Yang\thanks{\quad Equally Contribution}, Shiyu Fu\samethanks, Shuhong Chen, Xinyu Li and Ivan Marsic\\
  Multimedia Image Processing Lab\\
  Electrical and Computer Engineering Department\\
  Rutgers University, Piscataway, NJ, USA \\
  {\{yue.guapp, ky189, sf568, sc1624, Xinyu.li1118, marsic\}@rutgers.edu
  }}

\date{}

\begin{document}

\maketitle

\begin{abstract}
  Multimodal affective computing, learning to recognize and interpret human affect and subjective information from multiple data sources, is still challenging because: (i) it is hard to extract informative features to represent human affects from heterogeneous inputs; (ii) current fusion strategies only fuse different modalities at abstract levels, ignoring time-dependent interactions between modalities. Addressing such issues, we introduce a hierarchical multimodal architecture with attention and word-level fusion to classify utterance-level sentiment and emotion from text and audio data. Our introduced model outperforms state-of-the-art approaches on published datasets, and we demonstrate that our model's synchronized attention over modalities offers visual interpretability.
\end{abstract}

\section{Introduction}

With the recent rapid advancements in social media technology, affective computing is now a popular task in human-computer interaction. Sentiment analysis and emotion recognition, both of which require applying subjective human concepts for detection, can be treated as two affective computing subtasks on different levels \cite{poria2017review}. A variety of data sources, including voice, facial expression, gesture, and linguistic content have been employed in sentiment analysis and emotion recognition. In this paper, we focus on a multimodal structure to leverage the advantages of each data source. Specifically, given an utterance, we consider the linguistic content and acoustic characteristics together to recognize the opinion or emotion. Our work is important and useful because speech is the most basic and commonly used form of human expression.

A basic challenge in sentiment analysis and emotion recognition is filling the gap between extracted features and the actual affective states \cite{zhang2017learning}. The lack of high-level feature associations is a limitation of traditional approaches using low-level handcrafted features as representations \cite{seppi2008patterns, rozgic2012ensemble}. Recently, deep learning structures such as CNNs and LSTMs have been used to extract high-level features from text and audio \cite{eyben2010line, poria2015deep}. However, not all parts of the text and vocal signals contribute equally to the predictions. A specific word may change the entire sentimental state of text; a different vocal delivery may indicate inverse emotions despite having the same linguistic content. Recent approaches introduce attention mechanisms to focus the models on informative words \cite{yang2016hierarchical} and attentive audio frames \cite{mirsamadi2017automatic} for each individual modality. However, to our knowledge, there is no common multimodal structure with attention for utterance-level sentiment and emotion classification. To address such issue, we design a deep hierarchical multimodal architecture with an attention mechanism to classify utterance-level sentiments and emotions. It extracts high-level informative textual and acoustic features through individual bidirectional gated recurrent units (GRU) and uses a multi-level attention mechanism to select the informative features in both the text and audio module.

Another challenge is the fusion of cues from heterogeneous data. Most previous works focused on combining multimodal information at a holistic level, such as integrating independent predictions of each modality via algebraic rules \cite{wollmer2013youtube} or fusing the extracted modality-specific features from entire utterances \cite{poria2016convolutional}. They extract word-level features in a text branch, but process audio at the frame-level or utterance-level. These methods fail to properly learn the time-dependent interactions across modalities and restrict feature integration at timestamps due to the different time scales and formats of features of diverse modalities \cite{poria2017review}. However, to determine human meaning, it is critical to consider both the linguistic content of the word and how it is uttered. A loud pitch on different words may convey inverse emotions, such as the emphasis on ``hell'' for anger but indicating happy on ``great''. Synchronized attentive information across text and audio would then intuitively help recognize the sentiments and emotions. Therefore, we compute a forced alignment between text and audio for each word and propose three fusion approaches (horizontal, vertical, and fine-tuning attention fusion) to integrate both the feature representations and attention at the word-level.

We evaluated our model on four published sentiment and emotion datasets. Experimental results show that the proposed architecture outperforms state-of-the-art approaches. Our methods also allow for attention visualization, which can be used for interpreting the internal attention distribution for both single- and multi-modal systems. The contributions of this paper are: (i) a hierarchical multimodal structure with attention mechanism to learn informative features and high-level associations from both text and audio; (ii) three word-level fusion strategies to combine features and learn correlations in a common time scale across different modalities; (iii) word-level attention visualization to help human interpretation.

The paper is organized as follows: We list related work in section 2. Section 3 describes the proposed structure in detail. We present the experiments in section 4 and provide the result analysis in section 5. We discuss the limitations in section 6 and conclude with section 7.

\section{Related Work}

Despite the large body of research on audio-visual affective analysis, there is relatively little work on combining text data. Early work combined human transcribed lexical features and low-level handcrafted acoustic features using feature-level fusion \cite{forbes2004predicting, litman2004predicting}. Others used SVMs fed bag of words (BoW) and part of speech (POS) features in addition to low-level acoustic features \cite{seppi2008patterns, rozgic2012ensemble, savran2012combining, rosas2013multimodal, jin2015speech}. All of the above extracted low-level features from each modality separately. More recently, deep learning was used to extract higher-level multimodal features. Bidirectional LSTMs were used to learn long-range dependencies from low-level acoustic descriptors and derivations (LLDs) and visual features \cite{eyben2010line, wollmer2013youtube}. CNNs can extract both textual \cite{poria2015deep} and visual features \cite{poria2016convolutional} for multiple kernel learning of feature-fusion. Later, hierarchical LSTMs were used \cite{poria2017context}. A deep neural network was used for feature-level fusion in \cite{gu2018deep} and \cite{zadeh2017tensor} introduced a tensor fusion network to further improve the performance. A very recent work using word-level fusion was provided by \cite{chen2017multimodal}. The key differences between this work and the proposed architecture are: (i) we design a fine-tunable hierarchical attention structure to extract word-level features for each individual modality, rather than simply using the initialized textual embedding and extracted LLDs from COVAREP \cite{degottex2014covarep}; (ii) we propose diverse representation fusion strategies to combine both the word-level representations and attention weights, instead of using only word-level fusion; (iii) our model allows visualizing the attention distribution at both the individual modality and at fusion to help model interpretability.

Our architecture is inspired by the document classification hierarchical attention structure that works at both the sentence and word level \cite{yang2016hierarchical}. For audio, an attention-based BLSTM and CNN were applied to discovering emotion from frames \cite{huang2016attention, neumann2017attentive}. Frame-level weighted-pooling with local attention was shown to outperform frame-wise, final-frame, and frame-level mean-pooling for speech emotion recognition \cite{mirsamadi2017automatic}.

\section{Method}

\begin{figure}
	\centering
	\setlength{\abovecaptionskip}{1pt}
	\setlength{\belowcaptionskip}{-10pt}
	\includegraphics[width=1.0\linewidth]{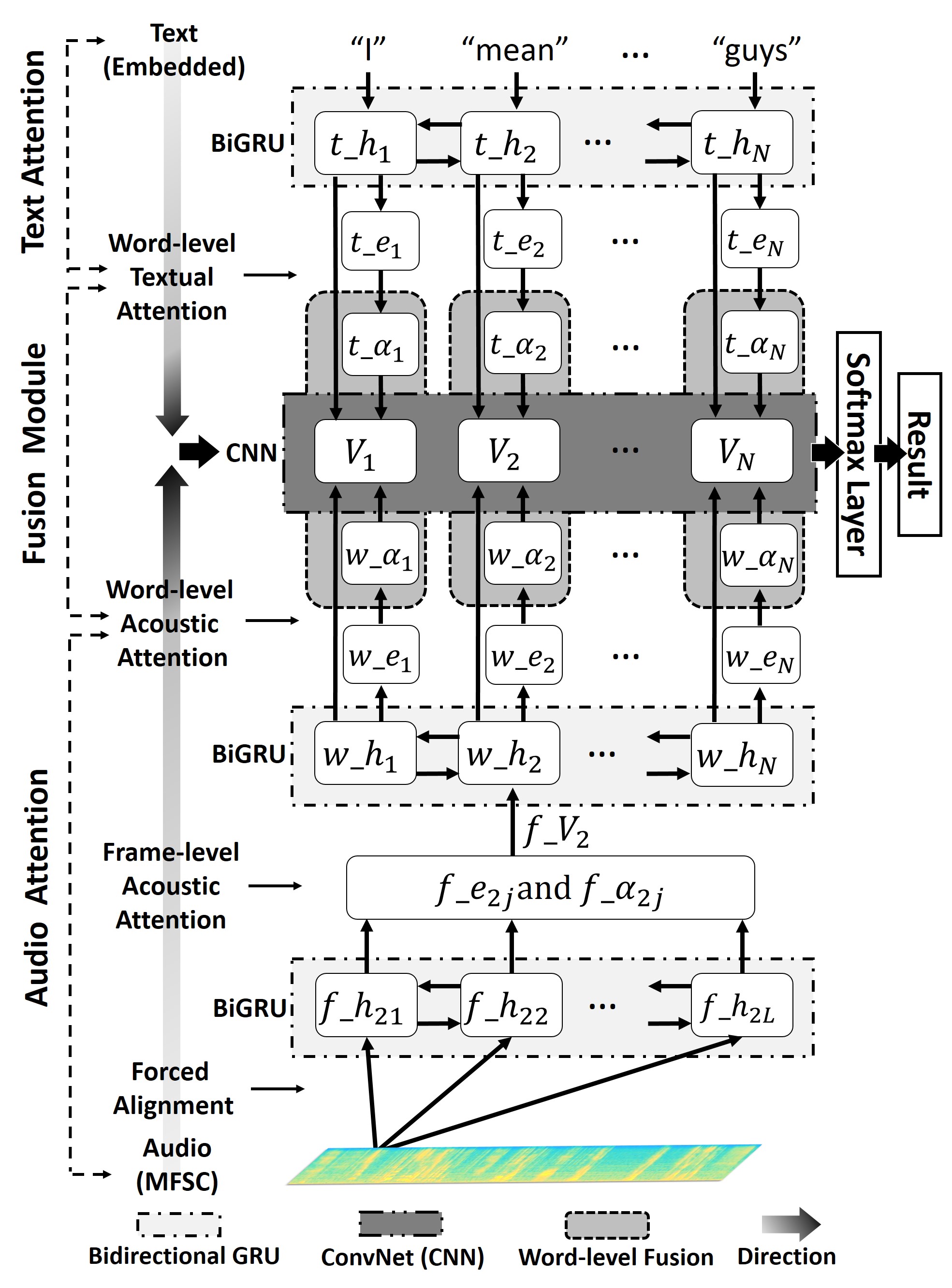}
	\caption{Overall Architecture}
	\label{fig:figure1}
\end{figure}

We introduce a multimodal hierarchical attention structure with word-level alignment for sentiment analysis and emotion recognition (Figure~\ref{fig:figure1}). The model consists of three major parts: text attention module, audio attention module, and word-level fusion module. We first make a forced alignment between the text and audio during preprocessing. Then, the text attention module and audio attention module extract the features from the corresponding inputs (shown in Algorithm 1). The word-level fusion module fuses the extracted feature vectors and makes the final prediction via a shared representation (shown in Algorithm 2).

\subsection{Forced Alignment and Preprocessing}
The forced alignment between the audio and text on the word-level prepares the different data for feature extraction. We align the data at the word-level because words are the basic unit in English for human speech comprehension. We used \textit{aeneas}\footnote{https://www.readbeyond.it/aeneas/} to determine the time interval for each word in the audio file based on the Sakoe-Chiba Band Dynamic Time Warping (DTW) algorithm \cite{sakoe1978dynamic}.

For the text input, we first embedded the words into 300-dimensional vectors by \textit{word2vec} \cite{mikolov2013distributed}, which gives us the best result compared to GloVe and LexVec. Unknown words were randomly initialized. Given a sentence $S$ with $N$ words, let $w_i$ represent the $i$th word. We embed the words through the \textit{word2vec} embedding matrix $W_e$ by:
\begin{equation}
T_i = W_ew_i,i \in [1,N]
\end{equation}
where $T_i$ is the embedded word vector.

For the audio input, we extracted Mel-frequency spectral coefficients (MFSCs) from raw audio signals as acoustic inputs for two reasons. Firstly, MFSCs maintain the locality of the data by preventing new bases of spectral energies resulting from discrete cosine transform in MFCCs extraction \cite{abdel2014convolutional}. Secondly, it has more dimensions in the frequency domain that aid learning in deep models \cite{gu2017speech}. We used 64 filter banks to extract the MFSCs for each audio frame to form the MFSCs map. To facilitate training, we only used static coefficients. Each word's MFSCs can be represented as a matrix with $64 \times n$ dimensions, where $n$ is the interval for the given word in frames. We zero-pad all intervals to the same length $L$, the maximum frame numbers of the word in the dataset. We did extract LLD features using OpenSmile \cite{eyben2010opensmile} software and combined them with the MFSCs during our training stage. However, we did not find an obvious performance improvement, especially for the sentiment analysis. Considering the training cost of the proposed hierarchical acoustic architecture, we decided the extra features were not worth the tradeoff. The output is a 3D MFSCs map with dimensions $[N, 64, L]$.

\begin{algorithm}[h]
\caption{FEATURE EXTRACTION}
\begin{algorithmic}[1]
\Procedure {FORCED ALIGNMENT}{}
\State Determine time interval of each word
\State \textbf{find} $w_i$ $\leftarrow$ $\rightarrow$ [$A_{ij}$], $j\in[1,L]$, $i\in[1,N]$
\EndProcedure
\Procedure{TEXT BRANCH}{}
\State Text Attention Module
\For {$i\in[1,N]$}
\State {$T_i \gets getEmbedded(w_i)$}
\State {$t\_h_i \gets bi\_GRU(T_i)$}
\State {$t\_e_i \gets getEnergies(t\_h_i)$}
\State {$t\_\alpha_i \gets getDistribution(t\_e_i)$}
\EndFor
\State return $t\_h_i$, $t\_\alpha_i$
\EndProcedure
\Procedure {AUDIO BRANCH}{}
\For {$i \in [1,N]$}
\State Frame-Level Attention Module
\For {$j \in [1,L]$}
\State {$f\_h_{ij} \gets bi\_GRU(A_{ij})$}
\State {$f\_e_{ij} \gets getEnergies(f\_h_{ij})$}
\State {$f\_\alpha_{ij} \gets getDistribution(f\_e_{ij})$}
\EndFor
\State {$f\_V_i \gets weightedSum(f\_\alpha_{ij}, f\_h_{ij})$}
\State Word-Level Attention Module
\State {$w\_h_i \gets bi\_GRU(f\_V_i)$}
\State {$w\_e_i \gets getEnergies(w\_h_i)$}
\State {$w\_\alpha_i \gets getDistribution(w\_e_i)$}
\EndFor
\State \textbf{return} {$w\_h_i, w\_\alpha_i$}
\EndProcedure
\end{algorithmic}
\end{algorithm}

\subsection{Text Attention Module}
To extract features from embedded text input at the word level, we first used bidirectional GRUs, which are able to capture the contextual information between words. It can be represented as:
\begin{equation}
t\_h^{\rightarrow}_i,t\_h^{\leftarrow}_i = bi\_GRU(T_i),i\in[1,N]
\label{equa:equa2}
\end{equation}
where $bi\_GRU$ is the bidirectional GRU, $t\_h^{\rightarrow}_i$ and $t\_h^{\leftarrow}_i$ denote respectively the forward and backward contextual state of the input text. We combined $t\_h^{\rightarrow}_i$ and $t\_h^{\leftarrow}_i$ as $t\_h_i$ to represent the feature vector for the $i$th word. We choose GRUs instead of LSTMs because our experiments show that LSTMs lead to similar performance (0.07\% higher accuracy) with around 25\% more trainable parameters.

To create an informative word representation, we adopted a word-level attention strategy that generates a one-dimensional vector denoting the importance for each word in a sequence \cite{yang2016hierarchical}. As defined by \cite{bahdanau2014neural}, we compute the textual attentive energies $t\_e_i$ and textual attention distribution $t\_\alpha_i$ by:
\begin{equation}
t\_e_i = tanh(W_tt\_h_i + b_t),i\in[1,N]
\label{equa:equa3}
\end{equation}
\begin{equation}
t\_\alpha_i = \frac{exp({t\_e_i}^{\top}v_t)}{{\sum}^N_{k=1}exp({t\_e_k}^{\top}v_t)}
\label{equa:equa4}
\end{equation}
where $W_t$ and $b_t$ are the trainable parameters and $v_t$ is a randomly-initialized word-level weight vector in the text branch. To learn the word-level interactions across modalities, we directly use the textual attention distribution $t\_\alpha_i$ and textual bidirectional contextual state $t\_h_i$ as the output to aid word-level fusion, which allows further computations between text and audio branch on both the contextual states and attention distributions.

\subsection{Audio Attention Module}
We designed a hierarchical attention model with frame-level acoustic attention and word-level attention for acoustic feature extraction.

\textbf{Frame-level Attention} captures the important MFSC frames from the given word to generate the word-level acoustic vector. Similar to the text attention module, we used a bidirectional GRU:
\begin{equation}
f\_h^{\rightarrow}_{ij},f\_h^{\leftarrow}_{ij} = bi\_GRU(A_{ij}),j\in[1,L]
\end{equation}
where $f\_h^{\rightarrow}_{ij}$ and $f\_h^{\leftarrow}_{ij}$ denote the forward and backward contextual states of acoustic frames. $A_{ij}$ denotes the MFSCs of the $j$th frame from the $i$th word, $i\in[1,N]$. $f\_h_{ij}$ represents the hidden state of the $j$th frame of the $i$th word, which consists of $f\_h^{\rightarrow}_{ij}$ and $f\_h^{\leftarrow}_{ij}$. We apply the same attention mechanism used for textual attention module to extract the informative frames using equation~\ref{equa:equa3} and \ref{equa:equa4}. As shown in Figure~\ref{fig:figure1}, the input of equation~\ref{equa:equa3} is $f\_h_{ij}$ and the output is the frame-level acoustic attentive energies $f\_e_{ij}$. We calculate the frame-level attention distribution $f\_\alpha_{ij}$ by using $f\_e_{ij}$ as the input for equation~\ref{equa:equa4}. We form the word-level acoustic vector $f\_V_i$ by taking a weighted sum of bidirectional contextual state $f\_h_{ij}$ of the frame and the corresponding frame-level attention distribution $f\_\alpha_{ij}$ Specifically,
\begin{equation}
f\_V_i = {\sum}_jf\_\alpha_{ij}f\_h_{ij}
\end{equation}

\begin{algorithm}[h]
\caption{FUSION}
\begin{algorithmic}[1]
\Procedure {FUSION BRANCH}{}
\State Horizontal Fusion (HF)
\For {$i \in [1,N]$}
\State {$t\_V_i \gets weighted(t\_\alpha_{i}, t\_h_{i})$}
\State {$w\_V_i \gets weighted(w\_\alpha_{i}, w\_h_{i})$}
\State {$V_i \gets dense([t\_V_i, w\_V_i])$}
\EndFor
\State Vertical Fusion (VF)
\For {$i \in [1,N]$}
\State {$h_i \gets dense([t\_h_{i}, w\_h_{i}])$}
\State {$s\_\alpha_{i} \gets average([t\_\alpha_{i}, w\_\alpha_{i}])$}
\State {$V_i \gets weighted(h_i, s\_\alpha_{i})$}
\EndFor
\State Fine-tuning Attention Fusion (FAF)
\For {$i \in [1,N]$}
\State {$u\_e_{i} \gets getEnergies(h_{i})$}
\State {$u\_\alpha_{i} \gets getDistribution(u\_e_{i},s\_\alpha_{i})$}
\State {$V_i \gets weighted(h_i, u\_\alpha_{i})$}
\EndFor
\State Decision Making
\State {$E \gets convNet(V_{1}, V_{2},...,V_{N})$}
\State \textbf{return} {E}
\EndProcedure
\end{algorithmic}
\end{algorithm}  

\textbf{Word-level Attention} aims to capture the word-level acoustic attention distribution $w\_\alpha_i$ based on formed word vector $f\_V_i$. We first used equation~\ref{equa:equa2} to generate the word-level acoustic contextual states $w\_h_i$, where the input is $f\_V_i$ and $w\_h_i = (w\_h^{\rightarrow}_i,w\_h^{\leftarrow}_i)$. Then, we compute the word-level acoustic attentive energies $w\_e_i$ via equation~\ref{equa:equa3} as the input for equation~\ref{equa:equa4}. The final output is an acoustic attention distribution $w\_\alpha_i$ from equation~\ref{equa:equa4} and acoustic bidirectional contextual state $w\_h_i$.

\begin{figure*}
	\centering
	\setlength{\abovecaptionskip}{1pt}
	\setlength{\belowcaptionskip}{-10pt}
	\includegraphics[width=1.0\linewidth]{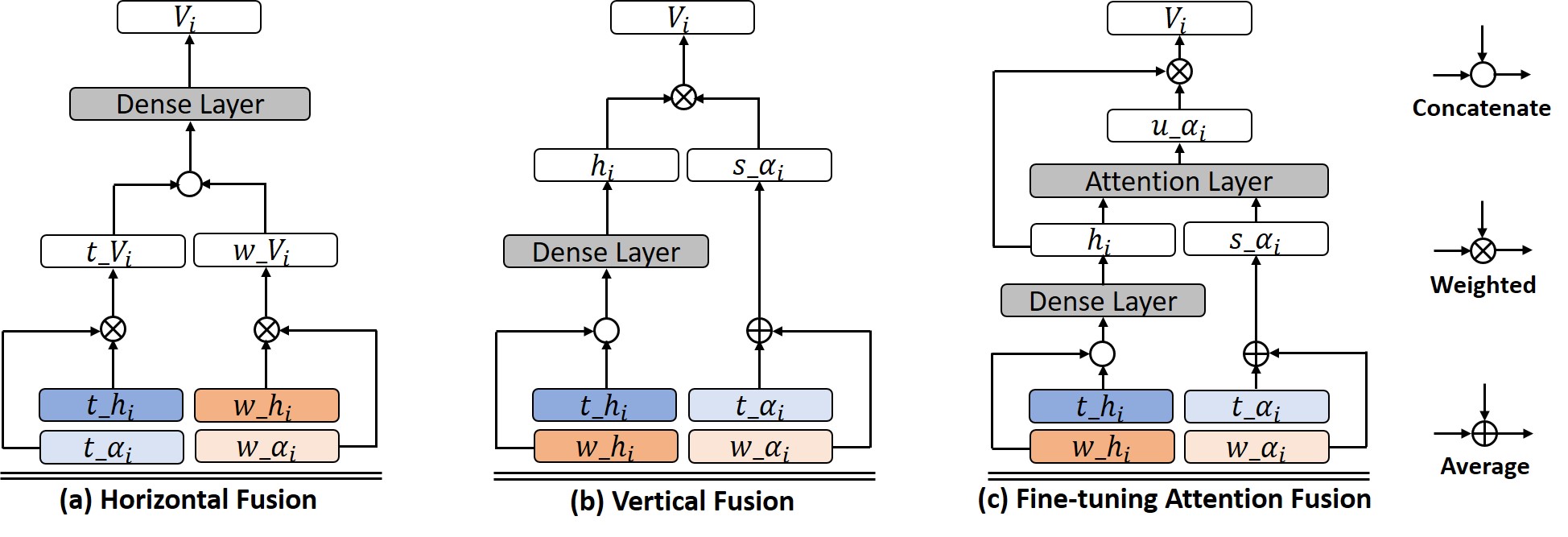}
	\caption{Fusion strategies. $t\_h_i$: word-level textual bidirectional state. $t\_\alpha_i$: word-level textual attention distribution. $w\_h_i$: word-level acoustic bidirectional state. $w\_\alpha_i$: word-level acoustic attention distribution. $s\_\alpha_i$: shared attention distribution. $u\_\alpha_i$: fine-tuning attention distribution. $V_i$: shared word-level representation.}
	\label{fig:figure2}
\end{figure*}

\subsection{Word-level Fusion Module}
Fusion is critical to leveraging multimodal features for decision-making. Simple feature concatenation without considering the time scales ignores the associations across modalities. We introduce word-level fusion capable of associating the text and audio at each word. We propose three fusion strategies (Figure~\ref{fig:figure2} and Algorithm 2): horizontal fusion, vertical fusion, and fine-tuning attention fusion. These methods allow easy synchronization between modalities, taking advantage of the attentive associations across text and audio, creating a shared high-level representation.

\textbf{Horizontal Fusion (HF)} provides the shared representation that contains both the textual and acoustic information for a given word (Figure~\ref{fig:figure2} (a)). The HF has two steps: (i) combining the bidirectional contextual states ($t\_h_i$ and $w\_h_i$ in Figure~\ref{fig:figure1}) and attention distributions for each branch ($t\_\alpha_i$ and $w\_\alpha_i$ in Figure~\ref{fig:figure1}) independently to form the word-level textual and acoustic representations. As shown in Figure~\ref{fig:figure2}, given the input ($t\_\alpha_i$, $t\_h_i$) and ($w\_\alpha_i$, $w\_h_i$), we first weighed each input branch by: 
\begin{equation}
t\_V_i = t\_\alpha_it\_h_i
\end{equation}
\begin{equation}
w\_V_i = w\_\alpha_iw\_h_i
\end{equation}
where $t\_V_i$ and $w\_V_i$ are word-level representations for text and audio branches, respectively; (ii) concatenating them into a single space and further applying a dense layer to create the shared context vector $V_i$, and $V_i = (t\_V_i,w\_V_i)$. The HF combines the unimodal contextual states and attention weights; there is no attention interaction between the text modality and audio modality.  The shared vectors retain the most significant characteristics from respective branches and encourages the decision making to focus on local informative features.

\textbf{Vertical Fusion (VF)} combines textual attentions and acoustic attentions at the word-level, using a shared attention distribution over both modalities instead of focusing on local informative representations (Figure~\ref{fig:figure2} (b)). The VF is computed in three steps: (i) using a dense layer after the concatenation of the word-level textual ($t\_h_i$) and acoustic ($w\_h_i$) bidirectional contextual states to form the shared contextual state $h_i$; (ii) averaging the textual ($t\_\alpha_i$) and acoustic ($w\_\alpha_i$) attentions for each word as the shared attention distribution $s\_\alpha_i$; (iii) computing the weight of $h_i$ and $s\_\alpha_i$ as final shared context vectors $V_i$, where $V_i = h_is\_\alpha_i$. Because the shared attention distribution ($s\_\alpha_i$) is based on averages of unimodal attentions, it is a joint attention of both textual and acoustic attentive information.

\textbf{Fine-tuning Attention Fusion (FAF)} preserves the original unimodal attentions and provides a fine-tuning attention for the final prediction (Figure\ref{fig:figure2} (c)). The averaging of attention weights in vertical fusion potentially limits the representational power. Addressing such issue, we propose a trainable attention layer to tune the shared attention in three steps: (i) computing the shared attention distribution $s\_\alpha_i$ and shared bidirectional contextual states $h_i$ separately using the same approach as in vertical fusion; (ii) applying attention fine-tuning:
\begin{equation}
u\_e_i = tanh(W_uh_i + b_u)
\label{equa:equa9}
\end{equation}
\begin{equation}
u\_\alpha_i = \frac{exp({u\_e_i}^{\top}v_u)}{{\sum}^N_{k=1}exp({u\_e_k}^{\top}v_u)} + s\_\alpha_i
\label{equa:equa10}
\end{equation}
where $W_u$, $b_u$, and $v_u$ are additional trainable parameters. The $u\_\alpha_i$ can be understood as the sum of the fine-tuning score and the original shared attention distribution $s\_\alpha_i$; (iii) calculating the weight of $u\_\alpha_i$ and $h_i$ to form the final shared context vector $V_i$.

\subsection{Decision Making}
The output of the fusion layer $V_i$ is the $i$th shared word-level vectors. To further make use of the combined features for classification, we applied a CNN structure with one convolutional layer and one max-pooling layer to extract the final representation from shared word-level vectors \cite{poria2016convolutional, wang2016select}. We set up various widths for the convolutional filters \cite{kim2014convolutional} and generated a feature map $c_k$ by:
\begin{equation}
f_i = tanh(W_cV_{i:i+k-1} + b_c)
\end{equation}
\begin{equation}
c_k = max\{f_1,f_2,...,f_N\}
\end{equation}
where $k$ is the width of the convolutional filters, $f_i$ represents the features from window $i$ to $i+k-1$. $W_c$ and $b_c$ are the trainable weights and biases. We get the final representation $c$ by concatenating all the feature maps. A softmax function is used for the final classification.

\section{Experiments}

\subsection{Datasets}
We evaluated our model on four published datasets: two multimodal sentiment datasets (MOSI and YouTube) and two multimodal emotion recognition datasets (IEMOCAP and EmotiW).

\textbf{MOSI} dataset is a multimodal sentiment intensity and subjectivity dataset consisting of 93 reviews with 2199 utterance segments \cite{zadeh2016mosi}. Each segment was labeled by five individual annotators between -3 (strong negative) to +3 (strong positive). We used binary labels based on the sign of the annotations' average.

\textbf{YouTube} dataset is an English multimodal dataset that contains 262 positive, 212 negative, and 133 neutral utterance-level clips provided by \cite{morency2011towards}. We only consider the positive and negative labels during our experiments.

\textbf{IEMOCAP} is a multimodal emotion dataset including visual, audio, and text data \cite{busso2008iemocap}. For each sentence, we used the label agreed on by the majority (at least two of the three annotators). In this study, we evaluate both the 4-catgeory (\textit{happy+excited}, \textit{sad}, \textit{anger}, and \textit{neutral}) and 5-catgeory(\textit{happy+excited}, \textit{sad}, \textit{anger}, \textit{neutral}, and \textit{frustration}) emotion classification problems. The final dataset consists of 586 \textit{happy}, 1005 \textit{excited}, 1054 \textit{sad}, 1076 \textit{anger}, 1677 \textit{neutral}, and 1806 \textit{frustration}.

\textbf{EmotiW}\footnote{https://cs.anu.edu.au/few/ChallengeDetails.html} is an audio-visual multimodal utterance-level emotion recognition dataset consist of video clips. To keep the consistency with the IEMOCAP dataset, we used four emotion categories as the final dataset including 150 \textit{happy}, 117 \textit{sad}, 133 \textit{anger}, and 144 \textit{neutral}. We used IBM Watson\footnote{https://www.ibm.com/watson/developercloud/speech-to-text/api/v1/} speech to text software to transcribe the audio data into text.

\begin{table*}
	\centering
	\scalebox{0.80}{
		\begin{tabular}{l|c|c|c|c|l|c|c|c|c}
        	\hline
            \multicolumn{5}{c|}{Sentiment Analysis (MOSI)} & \multicolumn{5}{c}{Emotion Recognition (IEMOCAP)} \\
			\hline
			Approach & Category & WA(\%) & UA(\%) & Weighted-F1 & Approach & Category & WA(\%) & UA(\%) & Weighted-F1 \\
			\hline
			BL-SVM* & \multicolumn{1}{c}{2-class} & \multicolumn{1}{c}{70.4} & \multicolumn{1}{c}{70.6} & 0.668 & SVM Trees & \multicolumn{1}{c}{4-class} & \multicolumn{1}{c}{67.4} & \multicolumn{1}{c}{67.4} & \multicolumn{1}{c}{-} \\
			LSTM-SVM* & \multicolumn{1}{c}{2-class} & \multicolumn{1}{c}{72.1} & \multicolumn{1}{c}{72.1} & 0.674 & GSV-e Vector & \multicolumn{1}{c}{4-class} & \multicolumn{1}{c}{63.2} & \multicolumn{1}{c}{62.3} & \multicolumn{1}{c}{-} \\
			C-MKL\textsubscript{1} & \multicolumn{1}{c}{2-class} & \multicolumn{1}{c}{73.6} & \multicolumn{1}{c}{-} & 0.752 & C-MKL\textsubscript{2} & \multicolumn{1}{c}{4-class} & \multicolumn{1}{c}{65.5} & \multicolumn{1}{c}{65.0} & \multicolumn{1}{c}{-}\\
			TFN & \multicolumn{1}{c}{2-class} & \multicolumn{1}{c}{75.2} & \multicolumn{1}{c}{-} & 0.760 & H-DMS & \multicolumn{1}{c}{5-class} & \multicolumn{1}{c}{60.4} & \multicolumn{1}{c}{60.2} & \multicolumn{1}{c}{0.594}\\ \cline{6-10}
			LSTM(A) & \multicolumn{1}{c}{2-class} & \multicolumn{1}{c}{73.5} & \multicolumn{1}{c}{-} & 0.703 & UL-Fusion* & \multicolumn{1}{c}{4-class} & \multicolumn{1}{c}{66.5} & \multicolumn{1}{c}{66.8} & \multicolumn{1}{c}{0.663} \\ \cline{1-5}
			UL-Fusion* & \multicolumn{1}{c}{2-class} & \multicolumn{1}{c}{72.5} & \multicolumn{1}{c}{72.5} & 0.730 & DL-Fusion* & \multicolumn{1}{c}{4-class} & \multicolumn{1}{c}{65.8} & \multicolumn{1}{c}{65.7} & \multicolumn{1}{c}{0.665} \\ \cline{6-10}
			DL-Fusion* & \multicolumn{1}{c}{2-class} & \multicolumn{1}{c}{71.8} & \multicolumn{1}{c}{71.8} & 0.720 & Ours-HF & \multicolumn{1}{c}{4-class} & \multicolumn{1}{c}{70.0} & \multicolumn{1}{c}{69.7} & \multicolumn{1}{c}{0.695} \\ \cline{1-5}
			Ours-HF & \multicolumn{1}{c}{2-class} & \multicolumn{1}{c}{74.1} & \multicolumn{1}{c}{74.4} & 0.744 & Ours-VF & \multicolumn{1}{c}{4-class} & \multicolumn{1}{c}{71.8} & \multicolumn{1}{c}{71.8} & \multicolumn{1}{c}{0.713} \\
			Ours-VF & \multicolumn{1}{c}{2-class} & \multicolumn{1}{c}{75.3} & \multicolumn{1}{c}{75.3} & 0.755 & Ours-FAF & \multicolumn{1}{c}{4-class} & \multicolumn{1}{c}{\textbf{72.7}} & \multicolumn{1}{c}{\textbf{72.7}} & \multicolumn{1}{c}{\textbf{0.726}}\\
			Ours-FAF & \multicolumn{1}{c}{2-class} & \multicolumn{1}{c}{\textbf{76.4}} & \multicolumn{1}{c}{\textbf{76.5}} & \textbf{0.768} & Ours-FAF & \multicolumn{1}{c}{5-class} & \multicolumn{1}{c}{\textbf{64.6}} & \multicolumn{1}{c}{\textbf{63.4}} & \multicolumn{1}{c}{\textbf{0.644}} \\
			\hline
		\end{tabular}}
	\caption{Comparison of models. \textit{WA} = weighted accuracy. \textit{UA} = unweighted accuracy. * denotes that we duplicated the method from cited research with the corresponding dataset in our experiment.}
	\label{results-comparison}
\end{table*}

\subsection{Baselines}
We compared the proposed architecture to published models. Because our model focuses on extracting sentiment and emotions from human speech, we only considered the audio and text branch applied in the previous studies.

\subsubsection{Sentiment Analysis Baselines}
\textbf{BL-SVM} extracts a bag-of-words as textual features and low-level descriptors as acoustic features. An SVM structure is used to classify the sentiments \cite{rosas2013multimodal}.

\textbf{LSTM-SVM} uses LLDs as acoustic features and bag-of-n-grams (BoNGs) as textual features. The final estimate is based on decision-level fusion of text and audio predictions \cite{wollmer2013youtube}.

\textbf{C-MKL\textsubscript{1}} uses a CNN structure to capture the textual features and fuses them via multiple kernel learning for sentiment analysis \cite{poria2015deep}.

\textbf{TFN} uses a tensor fusion network to extract interactions between different modality-specific features \cite{zadeh2017tensor}.

\textbf{LSTM(A)} introduces a word-level LSTM with temporal attention structure to predict sentiments on MOSI dataset \cite{chen2017multimodal}.

\subsubsection{Emotion Recognition Baselines}
\textbf{SVM Trees} extracts LLDs and handcrafted bag-of-words as features. The model automatically generates an ensemble of SVM trees for emotion classification \cite{rozgic2012ensemble}.

\textbf{GSV-eVector} generates new acoustic representations from selected LLDs using Gaussian Supervectors and extracts a set of weighed handcrafted textual features as an eVector. A linear kernel SVM is used as the final classifier \cite{jin2015speech}.

\textbf{C-MKL\textsubscript{2}} extracts textual features using a CNN and uses openSMILE to extract 6373 acoustic features. Multiple kernel learning is used as the final classifier \cite{poria2016convolutional}.

\textbf{H-DMS} uses a hybrid deep multimodal structure to extract both the text and audio emotional features. A deep neural network is used for feature-level fusion \cite{gu2018deep}.

\subsubsection{Fusion Baselines}
\textbf{Utterance-level Fusion (UL-Fusion)} focuses on fusing text and audio features from an entire utterance \cite{gu2017speech}. We simply concatenate the textual and acoustic representations into a joint feature representation. A softmax function is used for sentiment and emotion classification.

\textbf{Decision-level Fusion (DL-Fusion)} Inspired by \cite{wollmer2013youtube}, we extract textual and acoustic sentence representations individually and infer the results via two softmax classifiers, respectively. As suggested by W\"{o}llmer, we calculate a weighted sum of the text (1.2) result and audio (0.8) result as the final prediction.

\subsection{Model Training}
We implemented the model in Keras with Tensorflow as the backend. We set 100 as the dimension for each GRU, meaning the bidirectional GRU dimension is 200. For the decision making, we selected 2, 3, 4, and 5 as the filter width and apply 300 filters for each width. We used the rectified linear unit (ReLU) activation function and set 0.5 as the dropout rate. We also applied batch normalization functions between each layer to overcome internal covariate shift \cite{ioffe2015batch}. We first trained the text attention module and audio attention module individually. Then, we tuned the fusion network based on the word-level representation outputs from each fine-tuning module. For all training procedures, we set the learning rate to 0.001 and used Adam optimization and categorical cross-entropy loss. For all datasets, we considered the speakers independent and used an 80-20 training-testing split. We further separated 20\% from the training dataset for validation. We trained the model with 5-fold cross validation and used 8 as the mini batch size. We set the same amount of samples from each class to balance the training dataset during each iteration. 

\section{Result Analysis}

\subsection{Comparison with Baselines}
The experimental results of different datasets show that our proposed architecture achieves state-of-the-art performance in both sentiment analysis and emotion recognition (Table~\ref{results-comparison}). We re-implemented some published methods \cite{rosas2013multimodal, wollmer2013youtube} on MOSI to get baselines. 

For sentiment analysis, the proposed architecture with FAF strategy achieves 76.4\% weighted accuracy, which outperforms all the five baselines (Table~\ref{results-comparison}). The result demonstrates that the proposed hierarchical attention architecture and word-level fusion strategies indeed help improve the performance. There are several findings worth mentioning: (i) our model outperforms the baselines without using the low-level handcrafted acoustic features, indicating the sufficiency of MFSCs; (ii) the proposed approach achieves performance comparable to the model using text, audio, and visual data together \cite{zadeh2017tensor}. This demonstrates that the visual features do not contribute as much during the fusion and prediction on MOSI; (iii) we notice that \cite{poria2017context} reports better accuracy (79.3\%) on MOSI, but their model uses a set of utterances instead of a single utterance as input.

For emotion recognition, our model with FAF achieves 72.7\% accuracy, outperforming all the baselines. The result shows the proposed model brings a significant accuracy gain to emotion recognition, demonstrating the pros of the fine-tuning attention structure. It also shows that word-level attention indeed helps extract emotional features. Compared to C-MKL\textsubscript{2} and SVM Trees that require feature selection before fusion and prediction, our model does not need an additional architecture to select features. We further evaluated our models on 5 emotion categories, including frustration. Our model shows 4.2\% performance improvement over H-DMS and achieves 0.644 weighted-F1. As H-DMS only achieves 0.594 F1 and also uses low-level handcrafted features, our model is more robust and efficient.

From Table~\ref{results-comparison}, all the three proposed fusion strategies outperform UL-Fusion and DL-Fusion on both MOSI and IEMOCAP. Unlike utterance-level fusion that ignores the time-scale-sensitive associations across modalities, word-level fusion combines the modality-specific features for each word by aligning text and audio, allowing associative learning between the two modalities, similar to what humans do in natural conversation. The result indicates that the proposed methods improve the model performance by around 6\% accuracy. We also notice that the structure with FAF outperforms the HF and VF on both MOSI and IEMOCAP dataset, which demonstrates the effectiveness and importance of the FAF strategy.


\begin{figure*}
	\centering
	\setlength{\abovecaptionskip}{1pt}
	\setlength{\belowcaptionskip}{-10pt}
	\includegraphics[width=1.0\linewidth]{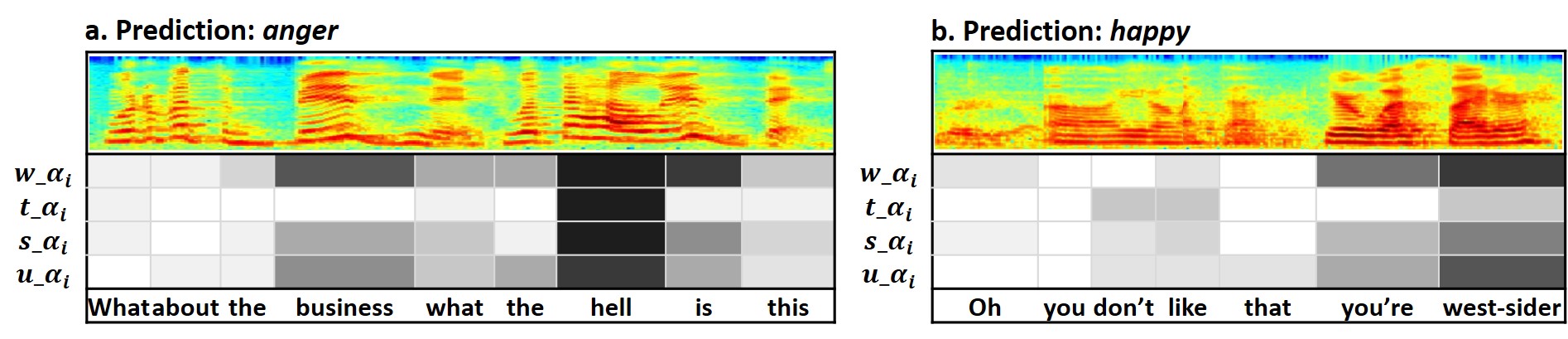}
	\caption{Attention visualization.}
	\label{fig:figure3}
\end{figure*}
\begin{table}
	\centering
	\setlength{\abovecaptionskip}{1pt}
	\begin{tabular}{c|c|c|c|c}
		\hline
		\multirow{2}{*}{Modality} & \multicolumn{2}{|c|}{MOSI} & \multicolumn{2}{|c}{IEMOCAP}\\
		\cline{2-5}
		& WA & F1 & WA & F1\\
		\hline
		T & \multicolumn{1}{c}{75.0} & 0.748 & \multicolumn{1}{c}{61.8} & 0.620\\
		A & \multicolumn{1}{c}{60.2} & 0.604 & \multicolumn{1}{c}{62.5} & 0.614\\
		T+A & \multicolumn{1}{c}{\textbf{76.4}} & \textbf{0.768} & \multicolumn{1}{c}{\textbf{72.7}} & \textbf{0.726}\\
		\hline
	\end{tabular}
	\caption{Accuracy (\%) and F1 score on text only (T), audio only (A), and multi-modality using FAF (T+A).}
	\label{table:table2}
\end{table}
\begin{table}
	\centering
	\setlength{\abovecaptionskip}{1pt}
	\setlength{\belowcaptionskip}{-10pt}
	\begin{tabular}{c|c|c|c|c}
		\hline
		\multirow{4}{*}{Approach} & \multicolumn{2}{|c|}{MOSI}  & \multicolumn{2}{|c}{IEMOCAP}\\
		& \multicolumn{2}{c|}{$\downarrow$} & \multicolumn{2}{c}{$\downarrow$}\\
		& \multicolumn{2}{c|}{YouTube} & \multicolumn{2}{c}{EmotiW}\\
		\cline{2-5}
		& WA & F1 & WA & F1\\
		\hline
		Ours-HF & \multicolumn{1}{c}{62.9} & 0.627 & \multicolumn{1}{c}{59.3} & 0.584\\
		Ours-VF & \multicolumn{1}{c}{64.7} & 0.643 & \multicolumn{1}{c}{60.8} & 0.591\\
		Ours-FAF & \multicolumn{1}{c}{\textbf{66.2}} &\textbf{ 0.665} & \multicolumn{1}{c}{\textbf{61.4}} & \textbf{0.608}\\
		\hline
	\end{tabular}
	\caption{Accuracy (\%) and F1 score for generalization testing.}
	\label{table:table3}
\end{table}

\subsection{Modality and Generalization Analysis}
From Table~\ref{table:table2}, we see that textual information dominates the sentiment prediction on MOSI and there is an only 1.4\% accuracy improvement from fusing text and audio. However, on IEMOCAP, audio-only outperforms text-only, but as expected, there is a significant performance improvement by combining textual and audio. The difference in modality performance might because of the more significant role vocal delivery plays in emotional expression than in sentimental expression.

We further tested the generalizability of the proposed model. For sentiment generalization testing, we trained the model on MOSI and tested on the YouTube dataset (Table~\ref{table:table3}), which achieves 66.2\% accuracy and 0.665 F1 scores. For emotion recognition generalization testing, we tested the model (trained on IEMOCAP) on EmotiW and achieves 61.4\% accuracy. The potential reasons that may influence the generalization are: (i) the biased labeling for different datasets (five annotators of MOSI vs one annotator of Youtube); (ii) incomplete utterance in YouTube dataset (such as ``about'', ``he'', etc.); (iii) without enough speech information (EmotiW is a wild audio-visual dataset that focuses on facial expression).

\subsection{Visualize Attentions}
Our model allows us to easily visualize the attention weights of text, audio, and fusion to better understand how the attention mechanism works. We introduce the emotional distribution visualizations for word-level acoustic attention ($w\_\alpha_i$), word-level textual attention ($t\_\alpha_i$), shared attention ($s\_\alpha_i$), and fine-tuning attention based on the FAF structure ($u\_\alpha_i$) for two example sentences (Figure~\ref{fig:figure3}). The color gradation represents the importance of the corresponding source data at the word-level.

Based on our visualization, the textual attention distribution ($t\_\alpha_i$) denotes the words that carry the most emotional significance, such as ``hell'' for \textit{anger} (Figure~\ref{fig:figure3} a). The textual attention shows that ``don't'', ``like'', and ``west-sider'' have similar weights in the \textit{happy} example (Figure~\ref{fig:figure3} b). It is hard to assign this sentence \textit{happy} given only the text attention. However, the acoustic attention focuses on ``you're'' and ``west-sider'', removing emphasis from ``don't'' and ``like''. The shared attention ($s\_\alpha_i$) and fine-tuning attention ($u\_\alpha_i$) successfully combine both textual and acoustic attentions and assign joint attention to the correct words, which demonstrates that the proposed method can capture emphasis from both modalities at the word-level.

\section{Discussion}
There are several limitations and potential solutions worth mentioning: (i) the proposed architecture uses both the audio and text data to analyze the sentiments and emotions. However, not all the data sources contain or provide textual information. Many audio-visual emotion clips only have acoustic and visual information. The proposed architecture is more related to spoken language analysis than predicting the sentiments or emotions based on human speech. Automatic speech recognition provides a potential solution for generating the textual information from vocal signals. (ii) The word alignment can be easily applied to human speech. However, it is difficult to align the visual information with text, especially if the text only describes the video or audio. Incorporating visual information into an aligning model like ours would be an interesting research topic. (iii) The limited amount of multimodal sentiment analysis and emotion recognition data is a key issue for current research, especially for deep models that require a large number of samples. Compared large unimodal sentiment analysis and emotion recognition datasets, the MOSI dataset only consists of 2199 sentence-level samples. In our experiments, the EmotiW and MOUD datasets could only be used for generalization analysis due to their small size. Larger and more general datasets are necessary for multimodal sentiment analysis and emotion recognition in the future.

\section{Conclusion}
In this paper, we proposed a deep multimodal architecture with hierarchical attention for sentiment and emotion classification. Our model aligned the text and audio at the word-level and applied attention distributions on textual word vectors, acoustic frame vectors, and acoustic word vectors. We introduced three fusion strategies with a CNN structure to combine word-level features to classify emotions. Our model outperforms the state-of-the-art methods and provides effective visualization of modality-specific features and fusion feature interpretation.

\section*{Acknowledgments}
We would like to thank the anonymous reviewers for their valuable comments and feedback. We thank the useful suggestions from Kaixiang Huang. This research was funded by the National Institutes of Health under Award Number R01LM011834. 

\bibliography{acl2018}
\bibliographystyle{acl_natbib}

\end{document}